\documentstyle[aaai,named]{article}
\nocopyright
\newcommand{\ifnmr}[1]{#1}

\newtheorem{example}{Example}
\newtheorem{theorem}{Theorem}

\newtheorem{corollary}{Corollary}
\newcommand{\fact}[2]{{\cal F}(#1,#2)}
\newcommand{\algorithmbox}[1]{
\ifnmr{\parbox{3.0in}{#1}}
}
\begin{document}
\title{Computing Circumscriptive Databases\\ by Integer Programming:
Revisited
\ifnmr{\\(Extended Abstract)}
} \author{
Ken Satoh, Hidenori Okamoto\\ Division of
Electronics and Information, Hokkaido University\\ N13W8 Kita-ku,
Sapporo, 060-8628, Japan\\ Email:ksatoh@db-ei.eng.hokudai.ac.jp\\
}
\date{}
\maketitle
\begin{abstract}
In this paper, we consider a method of computing minimal models in
circumscription using integer programming in propositional logic and
first-order logic with domain closure axioms and unique name axioms.
This kind of treatment is very important since this enable to apply
various technique developed in operations research to nonmonotonic
reasoning.

\cite{Nerode95} are the first to propose a method of
computing circumscription using integer programming. They claimed
their method was correct for circumscription with fixed predicate, but
we show that their method does not correctly reflect their claim. We
show a correct method of computing all the minimal models not only
with fixed predicates but also with varied predicates and we extend
our method to compute prioritized circumscription as well.
\end{abstract}
\section{Introduction}

In this paper, we discuss a method of computing circumscription using
integer programming used in operations research.
Circumscription~\cite{McCarthy86} has been proposed as a formalization
of nonmonotonic reasoning and intensively studied. However, like other
formalisms of nonmonotonic reasoning, it has a high complexity of
computation and many proposals are
made~\cite{Lifschitz85,Przymusinski89,Ginsberg89,Nerode95}.

\cite{Lifschitz85} gives a condition in which circumscriptive theory
is collapsed into the first-order logic.  \cite{Ginsberg89} and
\cite{Przymusinski89} give methods which use theorem prover
techniques.

\cite{Bell92,Bell96} and \cite{Nerode95} take different approach from the
above approaches.  Circumscription is restricted to a propositional
logic or a first-order sentences with domain closure axioms and unique
name axioms. Then, they translate axioms into inequality constraints
in integer programming and use a minimization of an objective function
which corresponds with minimized predicates and obtain all the minimal
models.  This kind of research is very important since it introduces an
usage of efficient method developed in operations research to
nonmonotonic reasoning.

In circumscription, there are three kinds of predicates; minimized
predicates, fixed predicates, and varied predicates. Minimized
predicates are subject to minimization whereas interpretation of fixed
predicates cannot be changed for minimization, but interpretation of
varied predicates can be changed if their change leads to further
minimization of minimized predicates. \cite{Bell92,Bell96} consider
minimization of all the predicates and \cite{Nerode95} claim that they
extend the method of \cite{Bell92,Bell96} so that their method is correct for
circumscription even including fixed predicates (but not including
varied predicates). However, we show that their claim is not correct.

Even if their claim were correct, circumscription without varied
predicates would have a serious drawback to apply circumscription to
commonsense reasoning as Etherington et al.~\cite{Etherington85} have
pointed out.

For example, consider the following axioms.

$bird\land \neg ab\supset fly.$

$bird.$

It seems that circumscribing $ab$ would yield $fly$. However, without
$fly$ varied, it is impossible to derive $fly$. This is because in
this circumscription without $fly$ varied, the interpretations are not
comparable each other if the interpretations of $fly$ are different.
There are three models of the above axioms, $I_1=\{bird, ab, fly\}$,
$I_2=\{bird, fly\}$, $I_3=\{bird, ab\}$\footnote{We represent an
interpretation as a set of true propositions in the interpretation.}.
In minimizing $ab$ without $fly$ varied, $I_2<I_1$ holds, but
$I_2<I_3$ does not hold since the interpretation of $fly$ in $I_2$ is
different from the interpretation of $fly$ in $I_3$.  So, minimal
models for this circumscription are $I_2$ and $I_3$, and therefore, we
cannot conclude $fly$.

If we let the interpretation of $fly$ be varied, then $I_2$ is the
only minimal model and therefore, we can conclude $fly$. Therefore,
usage of varied propositions is very important in commonsense
reasoning.

In this paper, we give a computing method of circumscription for a
propositional logic or a first-order logic with domain closure axioms
and unique name axioms. Our method can compute minimal models for this
class of axioms not only with fixed predicates, but also with varied
predicates. Moreover, \cite{Nerode95} gives a checking method of
circumscriptive entailment for a limited class of formulas, whereas we
give a complete checking method.  Then, we extend our method to apply
for prioritized circumscription as well.

\cite{Cadoli92} propose a method of eliminating varied predicates in
circumscription by translating inference problem of a formula under
circumscription with varied predicates and fixed predicates into
another inference problem under circumscription without varied
predicates nor fixed predicates. So, readers might think that methods
of \cite{Bell92,Bell96} which compute all the minimal models without
varied nor fixed propositions are sufficient for computing minimal
models. However, it is not clear how to apply the method proposed by
\cite{Cadoli92} to computing minimal models since the relationship
between a model of the original circumscription and a model of the
translated circumscription is not known.

\section{Preliminaries}
We restrict our attention to propositional circumscription.  For the
first-order case with domain closure axioms and unique name axioms, we
can translate each ground atom into a distinct proposition.

We assume that all propositional formulas are translated into a set of
clauses of the form $L_1\lor L_2\lor...\lor L_n$ where $L_i$ is a
positive literal $p_i$ or a negative literal $\neg p_i$.

We associate each propositional symbol $p$ with variable $X_p$ for 0-1
variable which represents the truth value of $p$; If $X_p=1$, $p$ is
true and if $X_p=0$, $p$ is false. 
We also use an interpretation $I$ to represent
a solution of the assignments to variables from integer
programming. If $p\in I$, it represents $X_p=1$ and if $p\not\in I$,
it represents $X_p=0$.

Let $F$ and $G$ be tuples of formulas, $\langle
F_1,F_2,...,F_n\rangle$ and $\langle G_1,G_2,...,G_n\rangle$.  We
define $F\leq G$ as $\bigwedge_{i=1}^{n} F_i\supset G_i$.
We define $F<G$  as $F\leq G$ and $G\not\leq F$, and $F\approx G$ as
$F\leq G$ and $G\leq F$.

Let $A$ be a conjunction of formulas and ${\cal P}$ be a set of
propositional symbols used in $A$.  We divide ${\cal P}$ into disjoint
three tuples of propositions $P,Z,Q$ which are called {\em minimized
propositions}, {\em varied propositions}, and {\em fixed
propositions}. 

Circumscription of $P$ for $A$ with $Z$ varied is defined as follows.

$Circum(A;P;Z)=$

$A(P,Z)\land \neg \exists p\exists z(A(p,z)\land p<P).$

For a model theory of circumscription, we define an order of
interpretations to minimize $P$ with $Z$ varied is defined as follows.
Let $I$ be an interpretation and $\Phi$ be a tuple of propositional
symbols.  We define $I[\Phi]$ as $\{p\in\Phi|I\models p\}$ or,
equivalently, $I\cap\Phi$.

Let $I_1$ and $I_2$ be interpretations.

$I_1\leq^{P;Z} I_2$ if
\begin{enumerate}
\item
$I_1[Q]=I_2[Q]$.
\item
$I_1[P]\subseteq I_2[P]$.
\end{enumerate}
We define $I_1<^{P;Z} I_2$ as $I_1\leq^{P;Z} I_2$ and
$I_2\not\leq^{P;Z} I_1$.  A minimal model $M$ of $A(P,Z)$ w.r.t. $P$
with $Z$ varied is defined as follows.
\begin{enumerate}
\item
$M$ is a model of $A(P,Z)$.
\item
There is no model $M'$ of $A(P,Z)$ such that $M'<^{P;Z}M$.
\end{enumerate}

According to~\cite{Lifschitz85}, $I$ is a minimal model of $A(P,Z)$
w.r.t. $P$ with $Z$ varied if and only if $I$ is a model of
$Circum(A;P;Z)$.

\section{Computing Minimal Models without Varied Propositions}
\label{fixedsection}

Let $A$ be a set of clauses. Then, a
set of inequalities, $Tr(A)$, translated from $A$ is defined as
follows.

$Tr(A)=$
\par\quad
$\{X_{p_1}+...+X_{p_n}+(1-X_{q_1})+...+(1-X_{q_m})\geq 1|$
\par\qquad
$p_1\lor...\lor p_n\lor \neg q_1\lor ...\lor \neg q_n\in A\}$

Let $Z$ be empty. Then, the algorithm proposed in \cite{Nerode95} is in
Figure~\ref{Nerodecircum}.
We adapt their algorithm for propositional circumscription.
The algorithm works as follows. We start with $Tr(A)$ as the initial
constraints and minimize an objective function corresponding with
minimized propositions under $Tr(A)$. If we do not obtain any solution,
we are done. Otherwise, we add a constraint $AC$ which excludes
non-minimal models larger than the obtained solution.

\begin{figure}
\framebox{\algorithmbox{
\begin{description}
\item[Step 1:] Let $AC:=\emptyset$ and $SS:=\emptyset$.
\item[Step 2:] Minimizing $\displaystyle\sum_{p\in P}X_p$ under $Tr(A)\cup AC$
using 0-1 integer programming.
\item[Step 3:] If there is no solution for the above minimization,
output $SS$
\item[Step 4:] Otherwise, 
\begin{enumerate}
\item
Let $M$ be a solution of the above minimization.
\item
Add $M[P]$ to $SS$.
\item
Add $\displaystyle\sum_{p\in M[P]} X_p\leq |M[P]|-1$ to $AC$.
\item
Go to Step 2.
\end{enumerate}
\end{description}
}}
\caption{\label{Nerodecircum} The algorithm of Nerode et al.}
\end{figure}

\cite{Nerode95} claims the following on the correctness and
completeness of the above algorithm.  
\vskip 6pt\noindent
{\bf Claim}\cite[Theorem 1]{Nerode95}
\begin{em}
Output $SS$ from the algorithm in Figure~\ref{Nerodecircum}
is equivalent to

$\{M[P]|M$ is a minimal model of $A(P)$ with respect to $P$ with
no propositions varied $\}.$
\end{em}
\vskip 6pt\noindent
Unfortunately, this claim is not correct in general as the following
example shows.
\begin{example}
\label{counterexample}
Let $A(ab)$ be the following set of clauses.
\vskip 8pt
$\neg bird\lor ab\lor fly.$

$bird.$
\vskip 8pt\noindent
Then, the minimal models of $A(ab)$ with respect to $\langle ab\rangle$
are $M_1=\{bird, fly\}$ and $M_2=\{bird, ab\}$. Note that $fly$ is a fixed
proposition and so, the two models are incomparable since
interpretations of $fly$ are different in these two models.

However, from the algorithm in Figure~\ref{Nerodecircum}, we cannot
obtain $M_2$ as follows.

$Tr(A)$ is 

$1-X_{bird}+X_{ab}+X_{fly}\geq 1$

$X_{bird}\geq 1$
\par\noindent
By minimizing $X_{ab}$ using 0-1 integer programming
under $Tr(A)$, we obtain a solution
$X_{ab}=0, X_{bird}=1, X_{fly}=1$ which corresponds with $M_1$.

Then, we add $M_1[\langle ab\rangle]=\emptyset$ to $SS$ and 
we add the following constraint to $AC$.

$0\leq -1$.

Obviously, we cannot get any further solution. This means
that we cannot obtain a minimal model $M_2$
\end{example}

Therefore, the above claim does not work in general if there is a
fixed proposition. Although their method is not correct with
circumscription with fixed propositions, we later show that their
method actually works for circumscription with varied propositions
without fixed propositions.

\begin{figure}
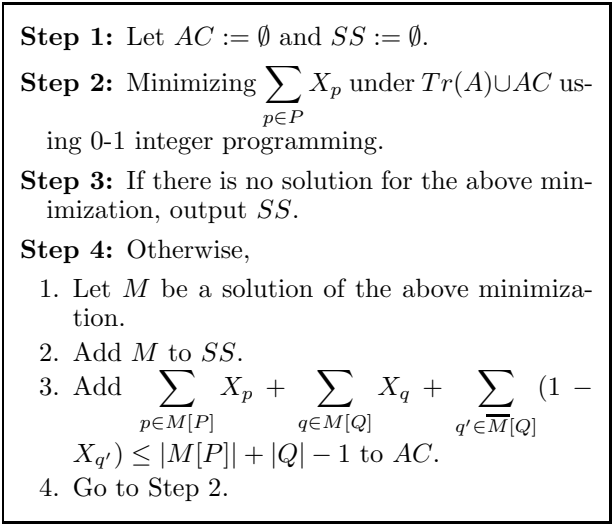

\framebox{\algorithmbox{
\begin{description}
\item[Step 1:] Let $AC:=\emptyset$ and $SS:=\emptyset$.
\item[Step 2:] Minimizing $\displaystyle\sum_{p\in P}X_p$ under $Tr(A)\cup AC$
using 0-1 integer programming.
\item[Step 3:] If there is no solution for the above minimization,
output $SS$.
\item[Step 4:] Otherwise, 
\begin{enumerate}
\item
Let $M$ be a solution of the above minimization.
\item
Add $M$ to $SS$.
\item
Add $\displaystyle\sum_{p\in M[P]} X_p
+\displaystyle\sum_{q\in M[Q]} X_q
+\displaystyle\sum_{q'\in \overline{M}[Q]}(1-X_{q'})
\leq |M[P]|+|Q|-1$ to $AC$.
\item
Go to Step 2.
\end{enumerate}
\end{description}
}}
\caption{\label{circwithfixed} Algorithm for circumscription with fixed propositions}
\end{figure}

Now, we give an algorithm which works correctly for circumscription
with fixed propositions in Figure~\ref{circwithfixed}.  Let $I$ be an
interpretation and $\Phi$ be a tuple of propositional symbols. We
define $\overline{I}[\Phi]$ used in Figure~\ref{circwithfixed} as
$\{p\in \Phi|I\not\models p\}$ or equivalently, $\Phi-I$.

\begin{theorem}
\label{fixedtheorem}
Output $SS$ from the algorithm in Figure~\ref{circwithfixed}
is equivalent to

$\{M|M$ is a minimal model of $A(P)$ with respect to $P$ with
no propositions varied $\}.$
\end{theorem}
{\bf Proof:} Let $\alpha$ be a formula which consists of logical
connectives and propositional symbols in $P$. Then, according
to~\cite{deKleer89}, $Circum(A;P)\models \alpha$ if and only if
$Circum(A\land (R\equiv \neg\cdot Q);P,Q,R)\models \alpha$ where $R$
is a tuple of new propositions not in $A$ and $\neg\cdot Q$ is
$\langle \neg q_1,...,\neg q_m\rangle$ for $Q=\langle
q_1,...,q_m\rangle$.  Then, we use the algorithm of~\cite{Bell92} to
minimize all propositions and replace every occurrence of variables
$X_{r_i}$ for a proposition
$r_i$ in $R$ by $1-X_{q_i}$.  $\Box$
\begin{example}
Let $A(ab)$ be the following set of clauses as in Example~\ref{counterexample}
\vskip 8pt
$\neg bird\lor ab\lor fly.$

$bird.$
\vskip 8pt\noindent
Then, the minimal models of $A(ab)$ with respect to $\langle ab\rangle$
are $M_1=\{bird, fly\}$ and $M_2=\{bird, ab\}$.

By minimizing $X_{ab}$ under $Tr(A)$, we obtain a solution $X_{ab}=0,
X_{bird}=1, X_{fly}=1$ which corresponds with a minimal model $M_1$.

Then, we add $M_1$ to $SS$ and we add the following constraint
to $AC$.

$X_{bird}+X_{fly}\leq 1$.

Then, minimizing $X_{ab}$ under $Tr(A)\cup AC$, we obtain
a solution $X_{ab}=1, X_{bird}=1, X_{fly}=0$
which corresponds with a minimal model $M_2$.

Then, we add $M_2$ to $SS$ and we add the following constraint
to $AC$.

$X_{ab}+X_{bird}+(1-X_{fly})\leq 2$

Then, minimizing $X_{ab}$ under $Tr(A)\cup AC$, we no longer obtain
any solution and therefore, $SS=\{\{bird,fly\},\{bird,ab\}\}$ is
obtained.
\end{example}

\section{Computing Minimal Models with Varied Propositions}
\label{withvaried}
\begin{figure}
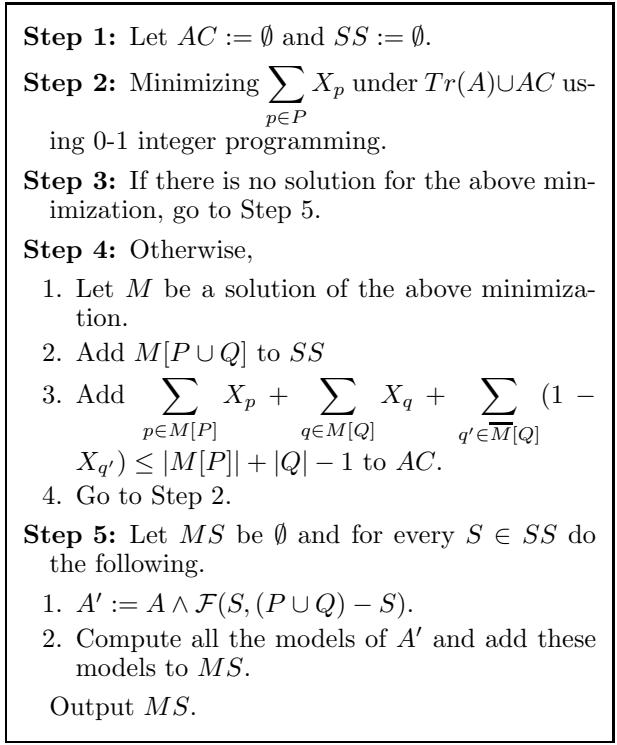

\framebox{\algorithmbox{
\begin{description}
\item[Step 1:] Let $AC:=\emptyset$ and $SS:=\emptyset$.
\item[Step 2:] Minimizing $\displaystyle\sum_{p\in P}X_p$ under $Tr(A)\cup AC$
using 0-1 integer programming.
\item[Step 3:] If there is no solution for the above minimization,
go to Step 5.
\item[Step 4:] Otherwise, 
\begin{enumerate}
\item
Let $M$ be a solution of the above minimization.
\item
Add $M[P\cup Q]$ to $SS$
\item
Add $\displaystyle\sum_{p\in M[P]} X_p
+\displaystyle\sum_{q\in M[Q]} X_q
+\displaystyle\sum_{q'\in \overline{M}[Q]}(1-X_{q'})
\leq |M[P]|+|Q|-1$ to $AC$.
\item
Go to Step 2.
\end{enumerate}
\item[Step 5:]
Let $MS$ be $\emptyset$ and for every $S\in SS$ do the following.
\begin{enumerate}
\item
$A':=A\land \fact{S}{(P\cup Q)-S}$.
\item
Compute all the models of $A'$ and add these models to $MS$.
\end{enumerate}
Output $MS$.
\end{description}
}}
\caption{\label{circwithvaried} Algorithm for circumscription with varied propositions}
\end{figure}

As shown in Introduction, we need varied proposition to perform
commonsense reasoning. We give a computation method of handling
varied propositions in Figure~\ref{circwithvaried}.

Let $F,G$ be disjoint sets of propositions.
We define $\fact{F}{G}$ as
$\displaystyle{\bigwedge_{f\in F} f\land \bigwedge_{f\in G}\neg f}$.

\begin{theorem}
Output $MS$ from the algorithm in Figure~\ref{circwithvaried}
is equivalent to

$\{M|M$ is a minimal model of $A(P,Z)$ with respect to $P$ with $Z$
varied $\}.$
\end{theorem}

\begin{example}
Let $A(ab)$ be the following set of clauses.
\vskip 8pt
$\neg bird\lor ab\lor fly.$

$bird.$
\vskip 8pt\noindent
Then, the minimal model of $A(ab)$ with respect to $\langle ab\rangle$
with $\langle fly\rangle$ varied
is $M_1=\{bird, ab\}$.

By minimizing $X_{ab}$ under $Tr(A)$, we obtain a solution where
$X_{ab}=0, X_{bird}=1$ and $X_{fly}=1$.  We add $M_1[\langle
ab\rangle\cup\langle bird\rangle]=\{bird\}$ to $SS$ and the following
constraint to $AC$.

$X_{bird}\leq 0$.

Obviously, there is no solution for $Tr(A)\cup AC$ and therefore,
$SS=\{\{bird\}\}$ is obtained.

Then, we add $\fact{\{bird\}}{\{ab,bird\}-\{bird\}}=bird\land \neg ab$
to $A$ to obtain $A'$ and compute all the models of $A'$.  We obtain
$MS=\{\{bird,fly\}\}$.
\end{example}

Actually, in the algorithm in Figure~\ref{circwithvaried}, if $Q$ is
empty and we output $SS$ at Step 3 instead of going to Step 5, then
this is equivalent to the algorithm of Nerode et al. In other words,
the correct claim for~\cite{Nerode95} is as follows.

\begin{corollary}
Let ${\cal P}$ be $P\cup Z$ and $Q$ be empty.
Final $SS$ in the algorithm in Figure~\ref{circwithvaried}
is equivalent to

$\{M[P]|M$ is a minimal model of $A(P,Z)$ with respect to $P$ with $Z$
varied $\}.$
\end{corollary}

If we only concern about circumscriptive entailment discussed
in~\cite{Nerode95}, that is, whether $Circum(A;P;Z)\models\alpha$ or
not, we do not need Step 5.  Instead, we check whether $A\land
\fact{S}{(P\cup Q)-S}\land\neg\alpha$ for every $S\in SS$ 
has any models or not. This can be done by checking whether
$Tr(A\land \fact{S}{(P\cup Q)-S}\land\neg
\alpha)$ does not have any solution when minimizing any arbitrary
objective function.  Note that in~\cite{Nerode95}, they use ``upper
and lower fringes'' to compute circumscriptive entailment for a
restricted class of formulas, but actually, such ``fringes'' are not
necessary.

\section{Computing Minimal Models in Prioritized Circumscription}
We firstly give a definition of prioritized circumscription.  We
divide a set of propositions into $n$ partitions and give an order
over partitions. Suppose that this is $P_1<P_2<...<P_n$. Intended
meaning of this order is that we firstly minimize $P_1$, then $P_2$ .... 
,then $P_n$.  Let $P$ and $Q$ be a tuple of propositions which have
orders $P_1<P_2<...<P_n$ and $Q_1<Q_2<...<Q_n$.  We define $P\preceq^i Q$ as
follows. If $i=1$, $P\preceq^i Q$ is $P_1\leq Q_1$ and if $i>1$,
$(\bigwedge_{j=1}^{i-1}P_j\approx Q_j)\supset P_i\leq Q_i$.  We define
$P\preceq Q$ as $\bigwedge_{i=1}^{n} P\preceq^{i} Q$ and $P\prec Q$ as
$P\preceq Q$ and $Q\not\preceq P$.

Prioritized circumscription of $P_1<P_2<...<P_n$ for $A$ with
$Z$ varied is defined as follows.

$Circum(A;P_1<P_2<...<P_n;Z)=$
\par\hfill
$A(P,Z)\land \neg \exists p\exists z(A(p,z)\land p\prec P).$

In a model theory of prioritized circumscription, we define
an order over interpretations as follows.

Let $I_1$ and $I_2$ be interpretations and let ${\cal P}$
consist of disjoint sets $P_1,P_2,...,P_n,Q,Z$.

$I_1\preceq^{P_1<P_2<...<P_n;Z} I_2$ if
\begin{enumerate}
\item
$I_1[Q]=I_2[Q]$.
\item
$I_1[P_1]\subseteq I_2[P_1]$.
\item
For every $i$, if for every $1\leq j\leq i-1$, $I_1[P_j]=I_2[P_j]$, then
$I_1[P_i]\subseteq I_2[P_i]$.
\end{enumerate}
We define $I_1\prec^{P_1<P_2<...<P_n;Z} I_2$ as
$I_1\preceq^{P_1<P_2<...<P_n;Z} I_2$ and
$I_2\not\preceq^{P_1<P_2<...<P_n;Z} I_1$.

A minimal model $M$ of $A(P,Z)$ w.r.t. $P_1<P_2<...<P_n$ with $Z$ varied
is defined as follows.
\begin{enumerate}
\item
$M$ is a model of $A(P,Z)$.
\item
There is no model $M'$ of $A(P,Z)$ such that
$M'\prec^{P_1<P_2<...<P_n;Z}M$.
\end{enumerate}

According to~\cite{Lifschitz85}, $I$ is a minimal model of
$A(P,Z)$ w.r.t. $P_1<P_2<...<P_n$ with $Z$ varied
iff $I$ is a model of $Circum(A;P_1<P_2<...<P_n;Z)$.

Similar to the problem in non-prioritized circumscription, the method
proposed in \cite{Nerode95} of computing prioritized circumscription
is correct if there are no fixed propositions.

To manipulate fixed propositions in prioritized circumscription, we
need the following theorem which is a generalization of the result of
~\cite{deKleer89}.
\begin{theorem}
\label{prioritizedtheorem}
Let a set of propositions
${\cal P}$ consist of disjoint sets $P_1,P_2,...,P_n,Q,Z$
and $P=P_1\cup P_2\cup...\cup P_n$
and $\alpha$ be a formula which consists of logical connectives
and propositional symbols in $P$. Then,

$Circum(A(P,Z);P_1>P_2>...>P_n;Z)\models\alpha$ if and only if
$Circum(A(P,Z)\land (R\equiv \neg\cdot Q);Q,R,P_1>P_2>...>P_n;Z)\models
\alpha$.
\end{theorem}

This theorem means that we can translate prioritized circumscription
with fixed propositions to prioritized circumscription without fixed
propositions. Moreover, we extend the method so that it is
applicable even if there are varied propositions.  We show the algorithm in
\ifnmr{Figure~\ref{prioritizedcircum}.}

\ifnmr{
\begin{figure}
\framebox{\algorithmbox{
\begin{description}
\item[Step 1:] Let $AC:=\emptyset$ and $SS:=\emptyset$.
\item[Step 2:] Minimizing $\displaystyle\sum_{p\in P_1}X_p$ under $Tr(A)\cup AC$
using 0-1 integer programming.
\item[Step 3:] If there is no solution for the above minimization,
go to Step 5.
\item[Step 4:] Otherwise, 
\begin{enumerate}
\item
Let $M$ be a solution of the above minimization.
\item
Add $M[P_1\cup Q]$ to $SS$
\item
Add $\displaystyle\sum_{p\in M[P_1]} X_p
+\displaystyle\sum_{q\in M[Q]} X_q
+\displaystyle\sum_{q'\in \overline{M}[Q]}(1-X_{q'})
\leq |M[P_1]|+|Q|-1$ to $AC$.
\item
Go to Step 2.
\end{enumerate}
\item[Step 5:]
For $i:=2$ to $n$ do the following.
\begin{enumerate}
\item
$SS':=\emptyset$.
\item
For every $S\in SS$ do
\begin{description}
\item[Step 5-1:] Let $AC:=\emptyset$.
\item[Step 5-2:] 
Minimizing $\displaystyle\sum_{p\in P_i}X_p$ under 
$Tr(A\land\fact{S}{(P_1\cup...\cup P_{i-1}\cup Q)-S})\cup AC$
using 0-1 integer programming.
\item[Step 5-3:] If there is no solution for the above minimization,
process the next $S$.
\item[Step 5-4:] Otherwise, 
\begin{enumerate}
\item
Let $M$ be a solution of the above minimization.
\item
Add $M[P_1\cup...\cup P_{i}\cup Q]$ to $SS'$.
\item
Add $\displaystyle\sum_{p\in M[P_i]} X_p \leq |M[P_i]|-1$ to $AC$.
\item
Go to Step 5-2.
\end{enumerate}
\end{description}
\item
$SS:=SS'$ and do the next iteration for $i$.
\end{enumerate}
If iteration stops then 
let $MS$ be $\emptyset$ and for every $S\in SS$ do the following.
\begin{enumerate}
\item Let $A':=A\land \fact{S}{(P_1\cup...\cup P_n\cup Q)-S}$.
\item Compute all the models of $A'$ and add these models to $MS$.
\end{enumerate}
Output $MS$.
\end{description}
}}
\caption{\label{prioritizedcircum} Algorithm for prioritized circumscription}
\end{figure}
}

\begin{theorem}
Output $MS$ from the algorithm in
\ifnmr{Figure~\ref{prioritizedcircum}}
is equivalent to

$\{M|M$ is a minimal model of $A(P,Z)$ with respect to $P_1<...<P_n$
with $Z$ varied$\}.$
\end{theorem}

\begin{example}
Consider the following axioms.

$ab_1\lor \neg fly$

$\neg bird\lor ab_2\lor fly.$

We compute minimal models of $Circum(A;\langle ab_2\rangle>\langle
ab_1\rangle;\langle fly\rangle)$ meaning that we minimize $ab_1$ and
$ab_2$ with $fly$ varied (and $bird$ fixed) and $ab_2$ is preferably
minimized than $ab_1$. The minimal models are $\{bird,fly,ab_1\}$ and
$\emptyset$. We have two minimal models since the interpretations of
$bird$ in these models are different from each other.

\begin{description}
\item[Step 1:] $AC:=\emptyset$ and $SS:=\emptyset$.
\item[Step 2(1):] Minimize $X_{ab_2}$ under the following constraints:
\par\quad
$X_{ab_1}+1-X_{fly}\geq 1$
\par\quad
$1-X_{bird}+X_{ab_2}+X_{fly}\geq 1$
\item[Step 3(1):]
Then, there are three solutions for this minimization:
\par\quad
$S_1=\{X_{ab_2}=0,X_{bird}=1,X_{fly}=1, X_{ab_1}=1\}$,
\par\quad
$S_2=\{X_{ab_2}=0,X_{bird}=0,X_{fly}=1, X_{ab_1}=1\}$,
\par\quad
$S_3=\{X_{ab_2}=0,X_{bird}=0,X_{fly}=0, X_{ab_1}=0\}$.
\item[Step 4(1):] Suppose that we obtain $S_1$.
\begin{enumerate}
\item $M_1=\{bird,fly,ab_1\}$
\item Add $M_1[\langle ab_2\rangle\cup \langle bird\rangle]=
\{bird\}$ to $SS$. ($SS$ becomes $\{\{bird\}\}$.)
\item Add $X_{bird}\leq 0$ to $AC$.
\end{enumerate}
\item[Step 2(2):]
Minimize $X_{ab_2}$ under new $AC$.
\item[Step 3(2):]
Then, there are two solutions for this minimization, $S_2$ and $S_3$.
\item[Step 4(2):] Suppose that we obtain $S_2$.
\begin{enumerate}
\item $M_2=\{fly,ab_1\}$.
\item Add $M_2[\langle ab_2\rangle\cup \langle bird\rangle]=\emptyset$
to $SS$. ($SS$ becomes $\{\{bird\},\emptyset\}$.)
\item Add $1-X_{bird}\leq 0$ to $AC$.
\end{enumerate}
\item[Step 2(3):]
Minimize $X_{ab_2}$ under new $AC$.
\item[Step 3(3):]
Then, we no longer obtain any solutions,
and go to Step 5.
\end{description}
\par\noindent
{\bf Step 5:}
\par\noindent
$i:=2$ and $SS':=\emptyset$.
\begin{enumerate}
\item $S:=\{bird\}$.
\begin{description}
\item[Step 5-1(1):] $AC:=\emptyset$.
\item[Step 5-2(1):] Minimize $X_{ab_1}$ under the following constraints:
\par\quad
$X_{ab_1}+1-X_{fly}\geq 1$
\par\quad
$1-X_{bird}+X_{ab_2}+X_{fly}\geq 1$
\par\quad
$X_{ab_2}\leq 0$
\par\quad
$X_{bird}\geq 1$

\item[Step 5-3(1):] Then, we obtain the solution $S_{1}$ again.
\item[Step 5-4(1):] $\;$
\begin{enumerate}
\item $M_{1}=\{bird, fly, ab_1\}$
\item Add $M_{1}[\langle ab_2\rangle\cup\langle ab_1\rangle\cup\langle bird\rangle]=\{ab_1,bird\}$ to $SS'$. ($SS'$ becomes $\{\{ab_1,bird\}\}$.)
\item Add $X_{ab_1}\leq 0$ to $AC$.
\end{enumerate}
\item[Step 5-2(2):] Minimize $X_{ab_1}$ under new $AC$.
\item[Step 5-3(2):] Then, we no longer obtain any solutions.
\end{description}
\item $S:=\emptyset$.
\begin{description}
\item[Step 5-1(1):] $AC:=\emptyset$.
\item[Step 5-2(1):] Minimize $X_{ab_1}$ under the following constraints:
\par\quad
$X_{ab_1}+1-X_{fly}\geq 1$
\par\quad
$1-X_{bird}+X_{ab_2}+X_{fly}\geq 1$
\par\quad
$X_{ab_2}\leq 0$
\par\quad
$X_{bird}\leq 0$

\item[Step 5-3(1):] Then, we obtain the solution $S_3$ only.
\item[Step 5-4(1):] $\;$
\begin{enumerate}
\item $M_3=\emptyset$
\item Add $M_3[\langle ab_2\rangle\cup\langle ab_1\rangle\cup\langle bird\rangle]
=\emptyset$ to $SS'$. ($SS'$ becomes $\{\{ab_1,bird\},\emptyset\}$.)
\item Add $0\leq -1$ to $AC$.
\end{enumerate}
\item[Step 5-2(2):] Minimize $X_{ab_1}$ under new $AC$.
\item[Step 5-3(2):] Then, we no longer obtain any solutions.
\end{description}
\end{enumerate}
Iteration stops and by calculation
of $MS$ from $SS'$, we obtain $\{\{bird, fly, ab_1\},\emptyset\}$.
\end{example}

We can also give a method of circumscriptive entailment
in prioritized circumscription as in ordinary circumscription.
After iteration stops, we check for every $A'\in
SS$, $A'\land\neg \alpha$ does not have any models to check
whether $\alpha$ is consequence of the prioritized circumscription
or not.

\section{Conclusion}
Contributions of this paper are as follows.

\begin{enumerate}
\item We correctly give the method of computing all the
models of circumscription not only with fixed propositions, but also
with varied propositions.
\item We give a complete method of computing circumscriptive entailment
for propositional logic.
\item We also extend the method of computing minimal models
to include varied propositions in prioritized circumscription.
\end{enumerate}
\par\noindent
{\bf Acknowledgements}
This research is partly supported by Grant-in-Aid for Scientific
Research (Project No. 11878067), The Ministry of Education, Japan.
We also thank the anonymous referees for comments on this paper.


\begin{thebibliography}{}
\bibitem[\protect\citeauthoryear{Bell  \bgroup et al.\egroup }{1992}]{Bell92}
Bell, C., Nerode, A., Ng, R., and Subrahmanian, V. S.
\newblock 1992.
\newblock Implementing Deductive Database by Linear Programming.
\newblock {\em Proc. of PODS92}, pages 283 -- 292.

\bibitem[\protect\citeauthoryear{Bell  \bgroup et al.\egroup }{1996}]{Bell96}
Bell, C., Nerode, A., Ng, R., and Subrahmanian, V. S.
\newblock 1996.
\newblock Implementing Deductive Database by Mixed Integer Programming.
\newblock {\em ACM Transactions on Database Systems}, {\bf 21}, pages 238 -- 269.

\bibitem[\protect\citeauthoryear{Cadoli  \bgroup et al.\egroup }{1992}]{Cadoli92}
Cadoli, M., Eiter, T., and Gottlob, G.
\newblock 1992
\newblock An Efficient Method for Eliminating Varying Predicates from a
Circumscription.
\newblock {\em Artificial Intelligence}, {\bf 54}, pages 397 -- 410.

\bibitem[\protect\citeauthoryear{de Kleer and Konolige}{1989}]{deKleer89}
de Kleer, J., and Konolige, K.
\newblock 1989.
\newblock Eliminating the Fixed Predicates from a Circumscription.
\newblock {\em Artificial Intelligence}, {\bf 39}, pages 391 -- 398.

\bibitem[\protect\citeauthoryear{Etherington \bgroup et al.\egroup }{1985}]{Etherington85}
Etherington, D., Mercer, R., and Reiter, R.
\newblock 1985.
\newblock On the Adequacy of Predicate Circumscription for Closed-world
Reasoning.
\newblock {\em Computational Intelligence}, {\bf 1}, pages 11 -- 15.

\bibitem[\protect\citeauthoryear{Ginsberg}{1989}]{Ginsberg89}
Ginsberg, M.
\newblock 1989.
\newblock Circumscriptive Theorem Prover.
\newblock {\em Artificial Intelligence}, {\bf 39}, pages 209 -- 230.

\bibitem[\protect\citeauthoryear{Lifschitz}{1985}]{Lifschitz85}
Lifschitz, V.
\newblock 1985.
\newblock Computing Circumscription.
\newblock {\em Proc. of IJCAI-85}, pages 121 -- 127.

\bibitem[\protect\citeauthoryear{McCarthy}{1986}]{McCarthy86}
McCarthy, J.
\newblock 1986.
\newblock Applications of Circumscription to Formalizing Common-Sense Knowledge.
\newblock {\em Artificial Intelligence}, {\bf 28}, pages 89 -- 116.

\bibitem[\protect\citeauthoryear{Nerode \bgroup et al.\egroup }{1995}]{Nerode95}
Nerode, A., Ng, R. T., and Subrahmanian, V. S.
\newblock 1995.
\newblock Computing Circumscriptive Databases, I. Theory and Algorithms.
\newblock {\em Information and Computation}, {\bf 116}, pages 58--80.

\bibitem[\protect\citeauthoryear{Przymusinski}{1989}]{Przymusinski89}
\newblock Przymusinski, T.
\newblock 1989.
\newblock An algorithm to compute circumscription.
\newblock {\em Artificial Intelligence}, {\bf 38}, pages 49 -- 73.

\end{thebibliography}
\end{document}